\documentclass[letterpaper, 10 pt, conference]{ieeeconf}  
\usepackage{graphicx,animate} 
\usepackage{hyperref}

\usepackage[colorinlistoftodos]{todonotes}
\usepackage{url}
\usepackage{multirow}
\usepackage{array}
\usepackage[verbose]{placeins}
\usepackage[backend=biber, style=numeric, sorting=ynt]{biblatex} 
\usepackage{booktabs}
\usepackage{cleveref}
\usepackage{adjustbox}
\usepackage{pifont}
\usepackage[font=footnotesize]{caption}
\IEEEoverridecommandlockouts                              

\title{\LARGE \bf
The Audio-Visual BatVision Dataset for Research on Sight and Sound
}

\author{Amandine Brunetto$^{1*}$,  Sascha Hornauer$^{1*}$, Stella X. Yu$^{2}$, Fabien Moutarde$^{1}$
\thanks{$^{1}$Center for Robotics, MINES Paris, Université PSL, Paris, France}%
\thanks{$^{2}$University of Michigan, Ann Arbor, United States of America}%
\thanks{$^{*}$Equal Contribution}%
}

\let\oldtwocolumn\twocolumn
\renewcommand\twocolumn[1][]{%
    \oldtwocolumn[{#1}{
    \begin{center}
           \includegraphics[width=\textwidth]{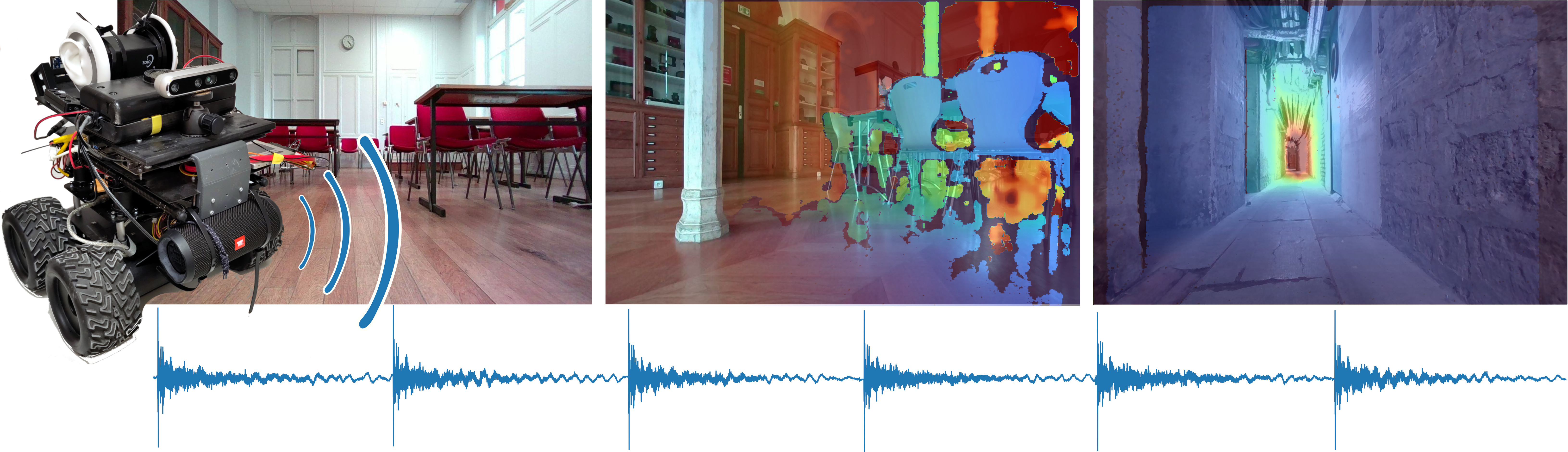}
           \captionof{figure}{The BatVision dataset contains large scale audio-visual data from a robots' perspective. For its creation, a robot traversed corridors, offices, lecture halls and driveways at a historic campus and modern office building like a bat, emitting \textit{chirping} sounds with a speaker. A binaural microphone recorded their echoes which carry rich scene information of objects, materials and layout. With this paper we provide echoes, camera images and depth maps, shown overlayed on typical scenes on the right. Echoes are shown under images. This dataset will help investigate fundamental questions on how sound interacts with spaces, how it can be harnessed for robotic navigation and what in general can be understood about a scene from how it sounds.}
           \label{fig:teaser_image}
        \end{center}
    }]
}

\begin{document}

\maketitle
\thispagestyle{empty}
\pagestyle{empty}

\begin{abstract}

Vision research showed remarkable success in understanding our world, propelled by datasets of images and videos. Sensor data from radar, LiDAR and cameras supports research in robotics and autonomous driving for at least a decade. However, while visual sensors may fail in some conditions, sound has recently shown potential to complement sensor data. Simulated room impulse responses (RIR) in 3D apartment-models became a benchmark dataset for the community, fostering a range of audiovisual research. In simulation, depth is predictable from sound, by learning bat-like perception with a neural network. Concurrently, the same was achieved in reality by using RGB-D images and echoes of \textit{chirping} sounds. Biomimicking bat perception is an exciting new direction but needs dedicated datasets to explore the potential. Therefore, we collected the BatVision dataset to provide large-scale echoes in  complex real-world scenes to the community. We equipped a robot with a speaker to emit \textit{chirps} and a binaural microphone to record their echoes. Synchronized RGB-D images from the same perspective provide visual labels of traversed spaces. We sampled modern US office spaces to historic French university grounds, indoor and outdoor with large architectural variety. This dataset will allow research on robot echolocation, general audio-visual tasks  and sound phænomena unavailable in simulated data. We show promising results for audio-only depth prediction and show how state-of-the-art work developed for simulated data can also succeed on our dataset. 
Project page: \url{https://amandinebtto.github.io/Batvision-Dataset/}

\end{abstract}


\section{Introduction}


Large-scale datasets propelled research in past decades, providing first static images and later an abundance of videos for tasks from object detection to activity recognition.

Sounds, correlated with visual data from their source, provide exploitable information about an action or context, often at marginal computational overhead.

A novel research direction aims to listen to the environment for improved task performance. Simulated room impulse response (RIR) datasets allow researchers to investigate the interaction of sounds with known space layouts for e.g. depth prediction, obstacle avoidance or to drive towards an alarm beyond the line of sight. They have been used successfully to predict 3D layouts from simulated chirps, similar to how bats find their prey. Robots mastering this echolocation could create instant maps beyond their immediate surroundings without LiDAR or cameras, overcoming their limitations in smoke \cite{zhang2015multi} and darkness.

\begin{figure*}[hbt!]
    \centering
     \smallskip
     \includegraphics[width=1.0\linewidth,angle=0,origin=c]{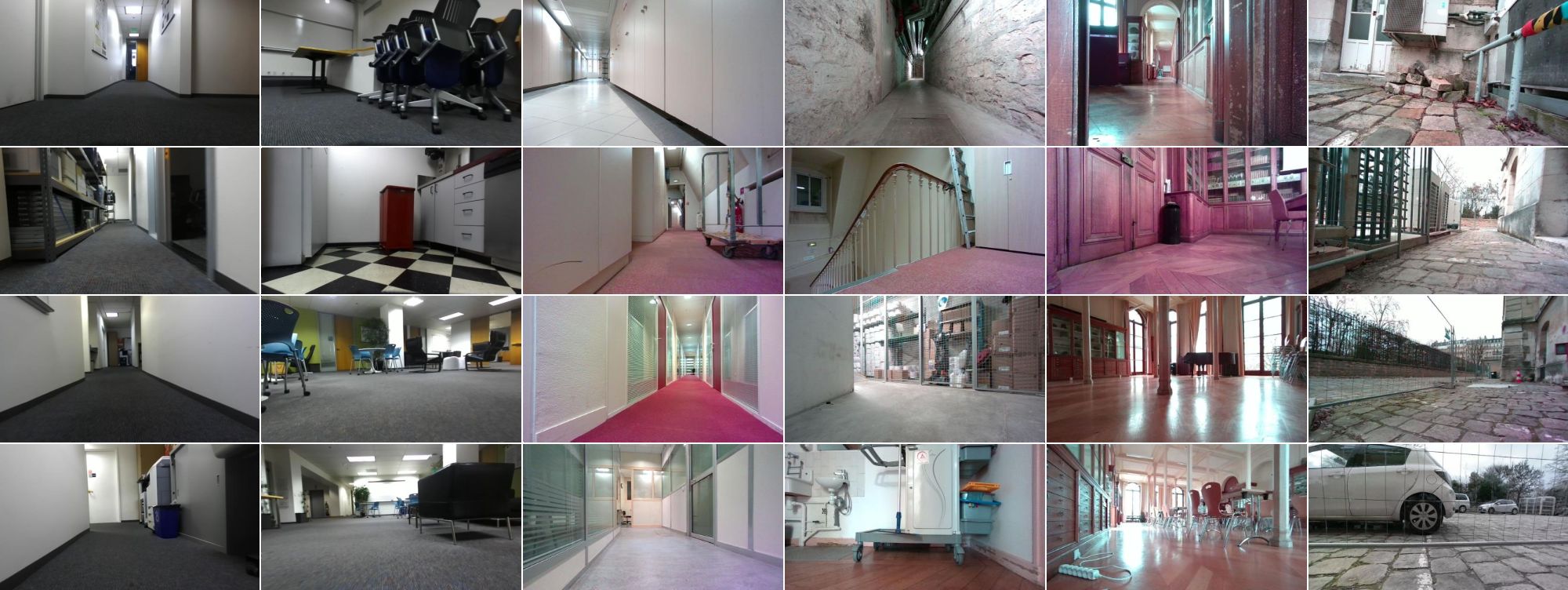}
        \caption{Example scenes from BV1 (left two columns) and BV2 (right four columns). BV1 contains typical office scenes with many corridors and some open spaces. BV2 columns show a wide variety of corridors, with and without carpet, maintenance areas, antique conference rooms and outdoor scenes.}     \label{fig:BVLocations}
 \end{figure*}

 \begin{figure}[tp]
    \centering
     \smallskip
     \includegraphics[width=1.0\linewidth,angle=0,origin=c]{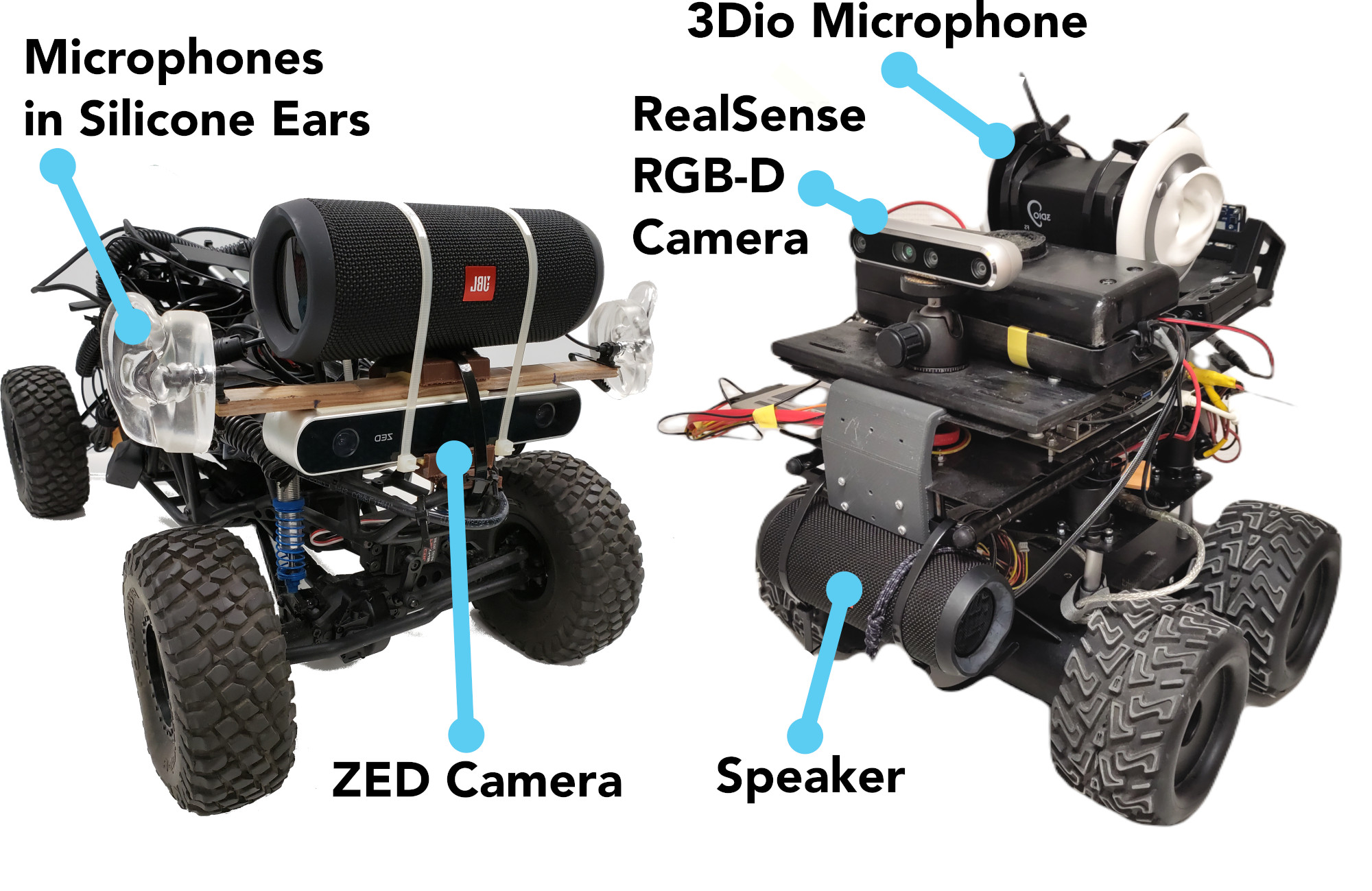}
        \caption{Recording robot used for BV1 at UC Berkeley (left) and BV2 at Mines Paris (right). Both record binaural audio and RGB-D images yet the hardware setup differs.}     \label{fig:robots}
 \end{figure}

While some datasets provide recorded RIRs within limitations, there exists no large-scale real dataset to investigate audio-visual 3D scene understanding for robots. This hinders domain adaptation from simulation as well as learning to be robust against or even exploit real world background sounds.

Bats are capable of navigating and localizing prey in flight using echolocation. We showed successfully how to adopt this feat for machine listening, using only audible echoes of chirps for depth prediction \cite{christensen2020batvision,christensen2020gccbatvision}. To understand the extend to which this method can complement failing vision sensors and even extend beyond the field of view, more real data in typical robotic scenarios is needed. We therefore present the BatVision Dataset providing publicly available large-scale real data of complex scenes for research on audio-visual 3D scene understanding and robotic echolocation.

\textbf{Sound-augmented Task Performance.} Sound arrives omnidirectional and provides rich information about the space it crosses. Due to the correlation of sound and vision it can provide complementary information when visual sensors fail. Cameras and LiDAR struggle with occlusions, reflections and smoke. Depth-from-stereo algorithms produce artifacts in low-light conditions and when objects lack textures. Combining audio-visual information can improve task performance with one modality providing labels for the other \autocite{schulz2021hearing,valverde2021there}. Biomimicking the echolocation principle can even improve 3D scene understanding using sound \autocite{christensen2020batvision,gao2020visualechoes,zhang2022stereo}. 

Audio-visual data suitable for these tasks is sparse because real-world recordings require complex setups and long sessions \cite{vcmejla2019mirage,hadad2014multichannel,9726801}. In consequence, researchers often create small datasets tailored to their tasks \cite{schulz2021hearing,majumder2023chat2map}). For research on robotic echolocation, only very few simple datasets exists.

\textbf{The BatVision Dataset.}
Motivated by the success of simulated datasets and the lack of real data for robotic echolocation and audio-visual scene understanding, we recorded the real-life BatVision dataset. We provide large-scale recordings from traversing real campus spaces with a robot, covering a wide range of materials, room shapes and objects, shown in \Cref{fig:BVLocations}. We provide synchronized RGB-D camera images and recorded echoes of \textit{chirps} sounds emitted with a forward facing speaker. Similar to RIRs, echoes can be used to infer the geometry of the room to allow the robot bat-like perception even in darkness, smoke and fog.

Data was collected at UC Berkeley (52,220 instances) and Mines Paris PSL (3,120 instances). Emitted \textit{chirp} signals range between 20\,Hz and 20\,kHz. RGB-D images were recorded with an Intel Realsense camera mounted on a robot (Mines Paris) and a ZED camera (UC Berkeley).

In the following we place the BatVision dataset among similar ones and argue its unique utility. We then detail the collection process, data and showcase performance of depth prediction methods trained on it. UC Berkeley data will be referred to as \textbf{BV1} and Mines Paris data as \textbf{BV2}.

\section{Related Work} \label{relatedworks}

Audio-visual data is available in many datasets, collected in reality or simulation, suitable for different tasks. However, to our knowledge, no existing dataset supports audio-visual 3D scene understanding with large-scale real data.

\textbf{Acoustic Room Impulse Response Datasets.} 
RIR datasets are used in room acoustics, speech and audio processing, sound localization and separation and virtual reality. RIRs account for the effect of room acoustics on audio signals and describe sound propagation. By convolving with RIRs, acoustic properties, such as reverb, can be transferred onto arbitrary \textit{dry} sounds.
Several datasets exist with RIRs, sampled in real spaces using a complex measurement routine.
 
The Acoustic Multichannel RIR Dataset  \cite{hadad2014multichannel} contains samples from one rectangular room. Panels, moved between recording sessions, emulate environments such as a small office, meeting or lecture room. The dEchorate dataset \cite{carlo2021dechorate} provides 1.8k sampled RIRs with annotated early echo timings and 3D positions of microphones and sound sources. For the MIRaGe dataset \cite{vcmejla2019mirage} 371k RIRs are measured in a dense grid of 4104 source positions in different rooms. 
\cite{kim2017acoustic} estimates RIRs from spherical camera images. RIRs were sampled as ground truth in living room style environments using a custom array of 48 microphones and a soundfield microphone. Object shapes are simple rectangles, aligned towards the main axis of the room.

While these datasets were sampled with high quality equipment and allow room geometry and RIRs estimation their environments are rectangular rooms with cuboid objects and simple materials. In contrast, for BatVision we sampled data in real public spaces, selected for their variety of materials, shape and architectural properties.

\textbf{Audio-Visual Simulation.}
Datasets containing simulated data are widely used, cost-effective, allow large scale generation, controlled conditions and easy data annotation \cite{shift2022},  \cite{gaidon2016virtual}). Simulation of sound propagation has been extensively used in games and AR applications. 

SoundSpaces 1.0 \cite{chen2020soundspaces} allows simulating sound propagation by providing RIR renderings built with bidirectional path-tracing in 3D-scanned apartment models.
They were generated for discreet positions and orientations in a grid in two 3D environments, Matterport3D \cite{chang2017matterport3d} and Replica\cite{straub2019replica}. SoundSpaces 2.0 \cite{chen2022soundspaces} provides continuous on-the-fly rendering and improved the sound propagation. 
It has since become a widely used benchmark showing impressive performance on a wide range of tasks  \cite{zhu2022beyond,gao2020visualechoes,zhu2022beyond,zhang2022stereo,parida2021beyond}. 

While it is a huge contribution to the community, bridging the gap between simulation and the real world stays a significant challenge. Models trained in simulation often overfit to characteristics of the simulator in surprising ways \autocite{kadian2020sim2real}. However, for audio-based 3D scene understanding there is no real-life dataset of a size, comparable to SoundSpaces which motivated our data collection.


\textbf{Sound Event Localization and Detection Datasets.} 
Sound event localization and detection (SELD) aims to infer the azimuth, elevation and distance of sounds relative to an observer, with or without additional classification. Datasets in the domain contain one or several, moving or static, clear or noisy sounds in different environments.
The TUT Acoustic Scenes 2016 dataset \cite{mesaros2016tut} consists of 15 real acoustic scenes, such as \textit{City center} or \textit{Metro station}, with annotated sound event classes, onset and offset times, and spatial information. The STARSS22 dataset \cite{politis2022starss22} is a collection of 22 real-world acoustic scene recordings with sound event annotations, captured with a high-resolution spherical microphone array. Annotations cover 13 classes and direction of arrivals.

For 3D scene understanding and especially reconstruction, SELD datasets often lack sufficient spatial ground truth information of the environments the sounds occur in. For the BatVision dataset forward facing camera images and depth information is provided to allow exploring the correlations in the audio and visual modality.

\textbf{Real-World Audio-Visual Datasets.} 
The huge EGO 4D dataset \cite{grauman2022ego4d} contains videos from the perspective of persons performing a wide range of activities. Parts contain audio and 3D meshes of the environment. It is an impressive effort and  contribution for many research areas such as activity recognition. In contrast, for BatVision we emit chirps into recorded scenes to allow inferring RIRs and finally the 3D scene. It is magnitudes smaller but better suited for our tasks.

\cite{owens2016visually} proposed “The Greatest Hits” dataset. By filming objects being struck with a drumstick, they aim to study physical interactions with a visual scene and synthesize plausible impact sounds. It includes 46,577 actions of hitting and scratching objects. The research is a step towards understanding the link between material, action and emitted sound. Similarly, with our recorded echoes we collect interactions between chirps and scene materials.

The authors in \cite{schulz2021hearing} predict approaching vehicles at blind intersections from sound before they enter the line-of-sight. They captured crossing vehicles at intersections with a custom microphone array and a front-facing camera mounted on a car. \cite{vasudevan2020semantic} recorded the "Omni Auditory Perception Dataset" standing next to streets with eight binaural microphones and a 360° camera. Pseudo ground truth depth is predicted from monocular images. The authors emphasize its mid-size and the great effort required to create it.

\cite{chen2021structure} introduced the "Quiet Campus Dataset" of ambient sounds from a variety of quiet indoor scenes. Paired with RGB-D images, they predict distances to walls from the whirling of a fan or noise coming through a window. While this unique idea shows impressive task performance with passive observations, we focus on active sounds which are humanly audible for improved performance.

Recording interactions of chirps with spaces has a recent history. The BatVision depth-from-binaural audio idea \cite{christensen2020batvision} inspired \cite{tracy2021catchatter} to collect similar data with added 360° LiDAR scans for depth and four ear-shaped microphones. With 5,000 samples the dataset falls between BV1 and BV2. 
\cite{majumder2023chat2map} predict an occupancy map from conversations in spaces. Beyond using SoundSpaces they also captured real data in a mock-up apartment, citing the lack of publicly available real world data. To compute RIRs they capture \textit{chirps} from a speaker with an Eigenmike. 
The "Studio Dataset" \cite{irie2022co} records 1478 samples of \textit{chirp} echoes with four microphones and RGB-D images with a RealSense camera. Everything is fixed into one metal frame facing in one direction. While also testing in SoundSpaces, their real-world recordings are done in one rectangular room showing simple cuboid objects.

The recording of so many datasets, tailored to specific needs show the potential impact that more and larger publicly available datasets can have. Presented tasks from depth prediction to occupancy mapping can be investigated with our BatVision dataset. We will describe in the following how and give details on the extend of our data collection.

\section{The Audio-Visual BatVision Dataset}

\newlength{\cw}
\setlength{\cw}{1.0\textwidth} 
\newcommand{\cmark}{\ding{51}}%
\newcommand{\xmark}{\ding{55}}%

{\setlength\tabcolsep{4pt}
\setlength{\extrarowheight}{2pt}
\begin{table*}[t]
    \centering
    \caption{Overview of Real-World Dataset. Mic. stands for microphone, A. for active  and P. for passive sound. Here, by \textit{chirp} we mean frequency swept signals with different range and parameters. Passive sound means no controlled sound is emitted during the recording. Instances are given either by their number or by the total amount of hours or videos recorded. }
    \resizebox{\textwidth}{!}{%
    \begin{tabular}{ ||m{0.05\cw}||m{0.10\cw}|m{0.13\cw}||m{0.15\cw}|m{0.12\cw}||m{0.10\cw}|m{0.10\cw}|m{0.10\cw}|| }%
        \hline
        \textbf{Dataset} & \textbf{\# Instances} & \textbf{Locations} & \textbf{Visual Labels} & \textbf{Perspective} & \textbf{Audio Labels} & \textbf{Audio Sensors} & \textbf{Sound Type} \\
        \hline
        \hline
        \textbf{BV1} & 52,220 & Hallways, open areas, conference rooms, \newline office spaces & Monocular RGB \newline Depth from stereo clipped at 12m & Robot & 72.5ms long binaural separated audio (44.1kHz) & 2 Hear-shaped Mic. & A. (\textit{chirp}) \\
        \hline
        \textbf{BV2} & 3,120 & Hallways,  outdoors, narrow underground corridors, conference rooms & Monocular RGB  \newline Depth from infrared at maximum 64m & Robot & 0.45s long binaural audio (44.1kHz) & Hear-shaped binaural Mic. & A. (\textit{chirp})\\
        \hline
        \multicolumn{8}{|l|}{\textbf{Acoustic Room Impulse Response Datasets}}\\\hline
        \cite{hadad2014multichannel} & 234 & Rectangular room. Panels emulate typical acoustic environments & \xmark & \xmark & 8 channels RIR (48kHz) & Mic. array & A. (\textit{chirp})\\
        \hline
        \cite{vcmejla2019mirage} & 371k & Rectangular room. Panels emulate typical acoustic environments & \xmark & \xmark & RIR (48kHz) & 6 Mic. arrays, \newline One additional mic. & A. (\textit{chirp}, white noise) \\
        \hline
        \cite{carlo2021dechorate} & $\approx$ 1.8k & Cubo\"{i}d room with different wall configurations & \xmark & \xmark & RIR (48kHz) & Mic. Array & A. (\textit{chirp}, white noise, anecho\"{i}c speech) \\
        \hline 
        \cite{kim2017acoustic} & ? & Living room, controlled acoustic environment & Stereo 360° RGB, \newline Depth from stereo & ? & RIR & Mic. circle & A. (\textit{chirp}) \\ 
        \hline 
        \multicolumn{8}{|l|}{\textbf{Sound Event Localization and Detection Datasets}}\\\hline
        \cite{mesaros2016tut} & $\approx$77k & 15 outdoor and indoor acoustic scenes & \xmark & \xmark & 30s long binaural audio (44.1kHz) annotated in sound classes & Binaural mic. & P. \\ 
        \hline 
        \cite{politis2022starss22} & $\approx$5h & Indoor rooms & 360° RGB & \xmark & 2x 4-channels spatial format audio (24kHz) annotated in 13 sound classes & Spherical Mic. Array (SMA) & P. \\
        \hline 
        \multicolumn{8}{|l|}{\textbf{Real-World Audio-Visuals Datasets}}\\\hline
        \cite{grauman2022ego4d} & 2,535h of videos with audio & Wide range of activities, cities and locations & RGB, 3D scans and other & Person & ? & Various & P. \\
        \hline 
        \cite{owens2016visually} & 46,577 & Indoor and outdoor scenes & RGB & Person & 35s long audio & Mic. attached to camera & A. (object stuck with drumstick) \\
        \hline 
        \cite{schulz2021hearing} & 411 video recordings & Urban environments & RGB & Car & 1s long audio (48kHz) & Mic. array & P. \\
        \hline 
        \cite{majumder2023chat2map} & ? & Mock-up apartment & RGB, \newline Depth from mono, \newline Occupancy map & Rig & RIR and conversation (16kHz) & SMA & A. (\textit{chirp}, speaker conversation) \\
        \hline 
        \cite{vasudevan2020semantic} & 54,250 video segments & Streets of a city covering 165 locations & 360° RGB & Rig & 3s long 8-channels audio (96kHz) & 4 hear-shaped binaural mic. & P. \\
        \hline
        \cite{irie2022co} & 1,478 & Near-rectangular reverberant studio & RGB, Depth & Rig & 1s long audio (16kHz) & 4 omnidirectional mic. & A. (\textit{chirp}) \\
        \hline 
        \cite{chen2021structure} & $\approx$15h & Classroom and hallways & RGB, Depth & Robot & 2 channels audio (16kHz) & Stereo mic. & P. \\
        \hline
    \end{tabular}}
    \label{tab:OverviewData}
\end{table*}
}

\subsection{Dataset Overview}
We collected the Audio-Visual BatVision Dataset at various locations at the UC Berkeley (ICSI), and Ecole des Mines Paris using small robots, carrying all sensors. The sites provide varied architectural styles, room shapes, materials and therefore acoustic impressions. 

We crossed lecture halls, corridors, offices and cobblestone paths recording while emitting humanly audible linear frequency sweep signals (\textit{chirps}) with a forward facing speaker (\Cref{fig:robots}). We recorded their echoes, forward facing camera images and provide depth-maps from the same perspective. \textit{Chirps} ascend from 20\,Hz to 20\,kHz in 3\,ms. Their use is motivated by their counterparts in nature, helping animals echolocalize, but they are also common in sound engineering to record RIRs \autocite{farina2000simultaneous}. Binaural microphones record audio at 44.1\,kHz with 24 bits to keep the full frequency range of the chirps. Each \textit{instance} is one \textit{chirp} synchronized with one 1280x720 RGB-D image.
An overview of this dataset and other real-world dataset is available in \Cref{tab:OverviewData}

\begin{figure}[t]
    \centering
    \smallskip
    \smallskip
    \includegraphics[width=1.0\columnwidth]{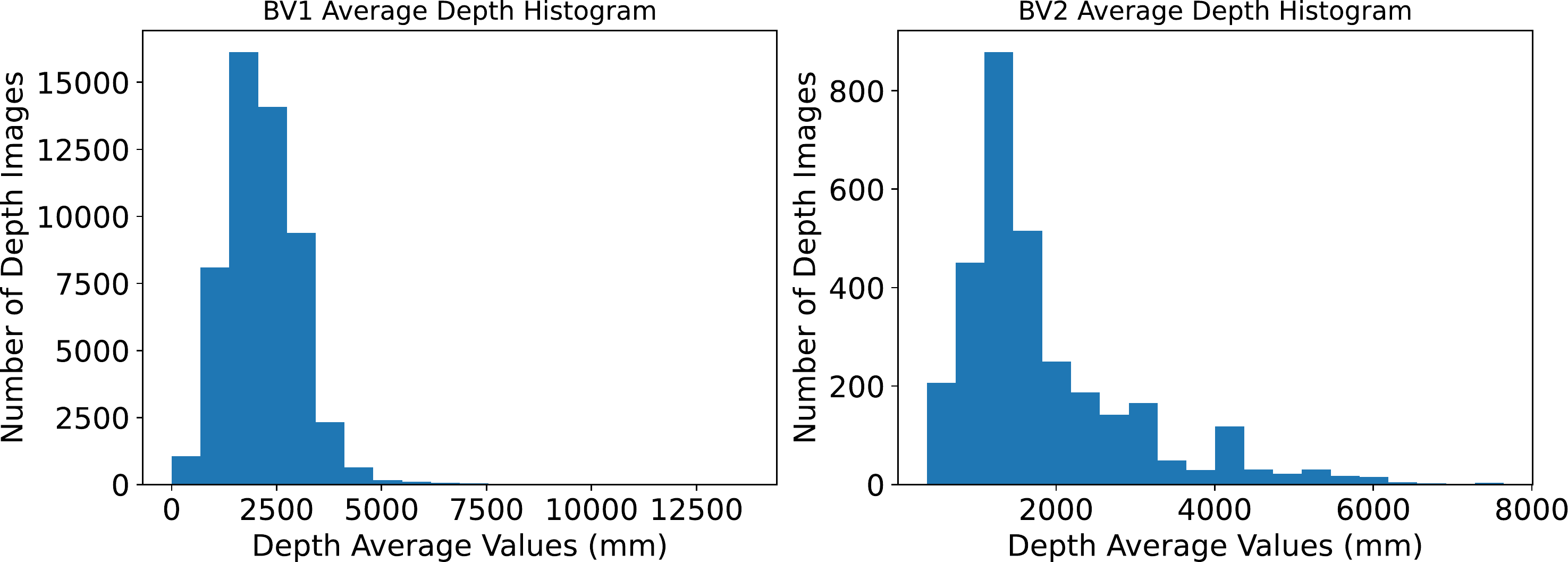}
    \caption{Histogram of average depth per instance. BV2 depth distribution is more long-tailed than BV1, which is consistent with the variety of data.}
    \label{fig:StatBV}
    \vspace{-0.5cm}
\end{figure}

\begin{figure}[t]
    \smallskip
    \smallskip
    \centering    \includegraphics[width=1.0\columnwidth]{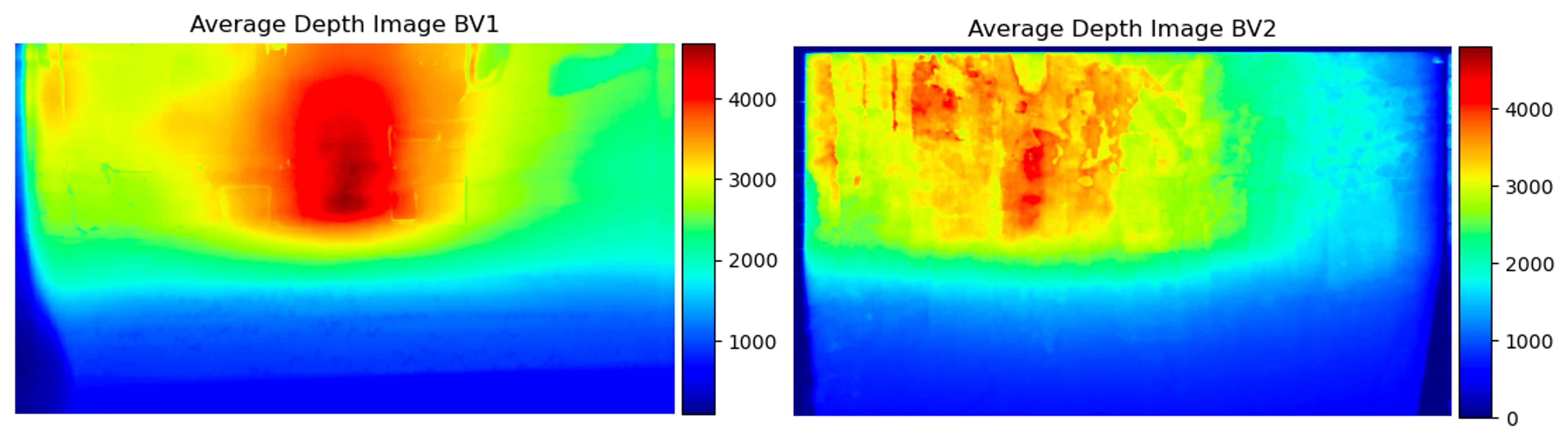}
    \caption{Average depth per pixel. In BV1 corridors are often centered, in BV2 depth distribution is more complex.}
    \label{fig:pixelwisedepth}
    \vspace{-0.5cm}
\end{figure}

The distribution of the collected depth can be seen at \Cref{fig:StatBV} and \Cref{fig:pixelwisedepth}. Comparing the average depth value per instances it shows BV2 is more long-tailed. BV1 depth values are clipped to the BV2 max depth value to remove outliers of the stereovision algorithm used. Pixel-wise average depth shows a more complex scene distribution in BV2.

\textbf{Data collected at the UC Berkeley (BV1).} 
In Berkeley, 52,220 instances were collected at two floors of an institute containing hallways, open areas, conference rooms and offices. 
At two distinct areas of one floor, 39,564 and 7,618 instances were collected for training and validation and 5,038 instances on another floor for testing. While similar, the floors' spatial layout, furniture, occupancy, and decorations are different, see \Cref{fig:BVLocations}. For details, please see \autocite{christensen2020batvision} showcasing the initial depth prediction-from-audio idea.

A JBL Flip4 Bluetooth speaker emitted chirps while two USB Lavalier MAONO AU-410 microphones, embedded into silicone ears, recorded their echoes. They were mounted 23.5cm apart while the speaker sat between (\Cref{fig:robots}). We excluded motor sounds for BV1 by pushing the robot on a trolley and for BV2 by stopping the remote-controlled robot. 

We used a ZED stereo camera to record images of the scene ahead and calculated depth maps with the  camera API. Depth-from-stereo fails for some pixel which are \textit{NAN}. We provide depth maps with RGB images from the left camera.

Designed for smaller spaces, audio recordings were cut at 72.5ms, including echoes from objects at 12\,m distance. This trades-off perception at a relevant distance while excluding later noisy reverberations. We also clip depth to 12\,m during training even though further distances are available.

\textbf{Data collected in Mines Paris (BV2).} 
In Ecole des Mines Paris, we collected 3,120 instances with large visual and acoustic variety (See \Cref{fig:BVLocations}).

We split data into sets of 1,911 train, 625 validation and 584 test instances. Given the multi-modal scene distribution we aim to balance task difficulty. We split instances by time of recording when moving through rooms, avoiding loops. That way, our incremental coverage of rooms will lead to poses being sufficiently separated in the sets. Even if we revisit parts of rooms we chose different trajectories thereby avoiding repeated poses. Tasks on this split will be harder than if instances were randomly assigned, which could result in neighbouring poses ending up in training and test. Simple interpolation will not yield best performance. Admittedly our split is easier than separating complete rooms as done for BV1. We observed outdoor reconstruction performance is acceptable but sub-par suggesting domain shift. If very different features, i.e. outdoors, carpeted rooms or stone corridors, are not in the training data further domain adaptation would be necessary which is left for future work.

We provide monocular RGB images and depth from an Intel RealSense D455 camera. The \textit{chirp-emitting} speaker is identical to BV1 but we recorded echoes with a binaural 3Dio Free Space microphone. A robot carried all hardware and an Nvidia Jetson TX-1 to run all recording software (see \Cref{fig:robots}). We excluded motor noise by switching between stopping and driving and filtered out instances while moving. 

Synchronization was achieved using ROS timestamps and \textit{chirps} detection with manual checks. Audio data instances were cut to be 0.45\,s long which includes echoes from objects up to 75\,m away. Because of larger spaces in BV2, we keep the long tail to sense far away reflection. The maximum depth value of the camera is $\approx$64m.

\subsection{Limitations}
Unlike in simulation, physically recording introduces a number of limitations. Echoes were recorded in real rooms next to noisy streets and with typical sounds of busy academic institutions. Some data may therefore contain audible noise, typical for the recording context. Models trained will need to learn to be robust to these noise profiles.

Some approaches predicting RIRs need independently changing emitter and microphone positions to record sounds on a direct path \autocite{carlo2021dechorate, luo2022learning}. In our data collection by driving, the robot carries all the equipment so the emitter is fixed close to the microphone and only the echoes change.


After filtering instances in motion few remain from the same pose. We kept these to allow learning a noise-model but they can be filtered by thresholding optical flow.

Collection with different hardware for BV1 and 2 leads to different formats. Audio instances collected at UC Berkeley are shorter. Some very bright images from Mines Paris shows slight purple discoloration due to an issues with the Intel RealSense D455 RGB-D camera. Finally, our consumer-grade speaker can not produce the full frequency spectrum.

\begin{figure}[tp]
    \centering
    \smallskip
    \includegraphics[width=1.0\columnwidth]{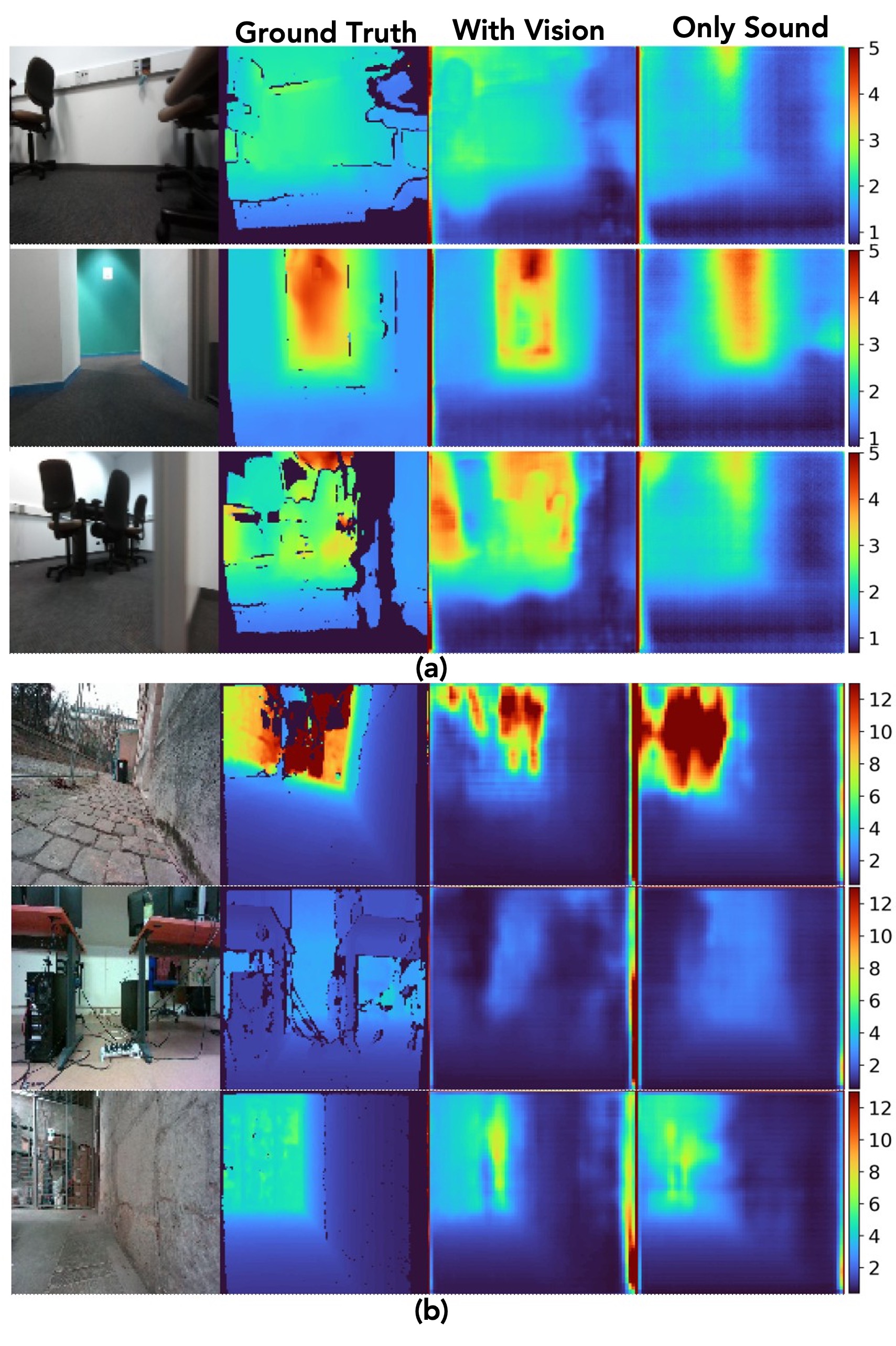}
    \caption{Test set results when training \textit{Beyond Image to Depth} on BatVision V1 (a) and BatVision V2 (b), depth in meters. Same hyperparameters as in simulation show well visible general layout and obstacles. Fifth row shows free space between two desks even based on audio-only. Fine structures such as cables are still hard to reconstruct.}
    \label{fig:bitd_examples}
\end{figure}

\section{Depth Prediction on BatVision Data}
The data can be used for depth-prediction from audio-visual data with approaches developed for real or simulated data. For illustration we trained a U-Net audio-only depth prediction baseline and compared with one state-of-the-art audio-visual approach developed by its authors in simulation.

\begin{figure*}[tp]
    \centering
    \smallskip
    \smallskip    \includegraphics[width=0.92\textwidth]{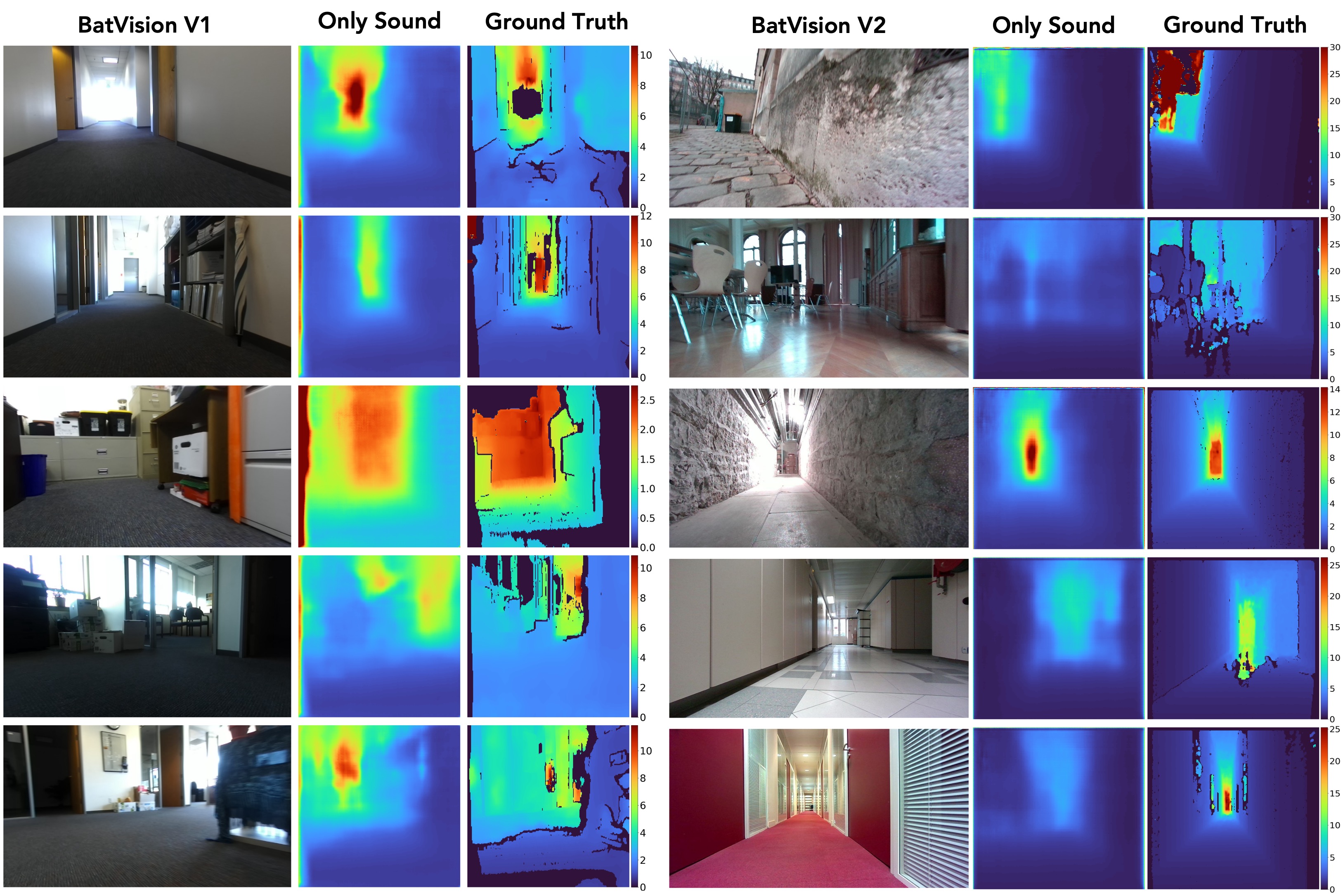}
    \caption{U-Net Baseline depth predictions on BatVision V1 (left) and BatVision V2 (right) ('\textit{turbo}' colormap). Columns are f.l.t.r. RGB image, depth prediction and ground truth depth. Units in meters. Left, beyond showing correct structures, results improved quantitatively in this larger dataset. Right, obstacles like chairs are reconstructed and corridor layouts are correct, even though wall materials, and hence echoes, vary strongly from carpet to stone. }
    \label{fig:UnetResults}
    \vspace{-0.5cm}
\end{figure*}

\textbf{Beyond Image to Depth.}
Recent work improves audio-visual depth prediction using a pre-trained material classifier to decide which modality to pay attention to. The authors \autocite{parida2021beyond} extract latent material features from SoundSpaces 1.0 camera images and generate attention maps for the visual and audio-stream in the network. They convolve a \textit{dry-chirp} sound with SoundSpace RIRs to simulate emitting it into spaces from different poses. Captured echoes and RGB-D images make up their dataset with which they predict depth from RGB images and audio.

Training their code on the BatVision dataset shows similar performance as on SoundSpaces (see \Cref{tab:bitd_table}). Apart from changing the spectrogram resolution, no further adjustments were needed when switching to real data. We ablate results in a study using the audio signal only by setting the RGB images to zero. That way we keep the architecture unchanged.
{\setlength\tabcolsep{3pt}
\begin{table}[hbt!]
    \centering
    \caption{Depth prediction results from Beyond Image to Depth \autocite{parida2021beyond}, trained on BatVision and simulated data (Replica and Matterport). RMSE unit is meters. Similar V1 and Matterport results suggest comparable difficulty of tasks. Our simple U-Net baseline slightly outperforms \autocite{parida2021beyond} when using audio-only (AO).}
    \resizebox{0.98\columnwidth}{!}{%
    \begin{tabular}{p{0.26\columnwidth}p{0.10\columnwidth}p{0.09\columnwidth}p{0.09\columnwidth}p{0.09\columnwidth}p{0.09\columnwidth}p{0.09\columnwidth}}\toprule
     & RMSE$\downarrow$ & REL$\downarrow$ & log10$\downarrow$ & $\delta_{1.25}\uparrow$ & $\delta_{1.25^2}\uparrow$ & $\delta_{1.25^3}\uparrow$ \\\midrule
    Replica \autocite{parida2021beyond} &\textbf{0.249}&\textbf{0.118}&\textbf{0.046}&\textbf{0.869}&\textbf{0.943}&\textbf{0.970} \\
    Matterport3D \autocite{parida2021beyond} & 0.950&0.175&0.079&0.733&0.886&0.948\\
    BV1 & \textbf{0.901} & \textbf{0.234}&\textbf{0.097}&\textbf{0.688}&\textbf{0.888}&\textbf{0.942} \\
    BV2 & 2.286 & 0.323 & 0.119 & 0.647 & 0.834 & 0.901 \\
    BV1 AO &1.350&0.453&0.159&0.441&0.707&0.843 \\
    BV2 AO & 2.878 & 0.521 & 0.197 & 0.430 & 0.629 & 0.765 \\\midrule
    U-Net BV1 AO& 1.336 & 0.361 & 0.147 & 0.508 & 0.738 & 0.856 \\
    U-Net BV2 AO& 2.676 & 0.432 & 0.160 & 0.497 & 0.717 & 0.835 \\
    \bottomrule
    \end{tabular}}
    
    \label{tab:bitd_table}
\end{table}}

Qualitatively, fine structures are harder to predict which may stem from our coarser ground truth depth compared to simulation (See \Cref{fig:bitd_examples}). Measuring depth in reality, either from stereo (ZED camera) or active stereo (RealSense camera), limits the accuracy depending on the object distance.

Comparable results show that an approach developed for simulated data can be adapted and works with the same hyperparameters on our real dataset. 

\begin{figure}[h!tb]
    \centering
    \includegraphics[width=1.0\columnwidth]{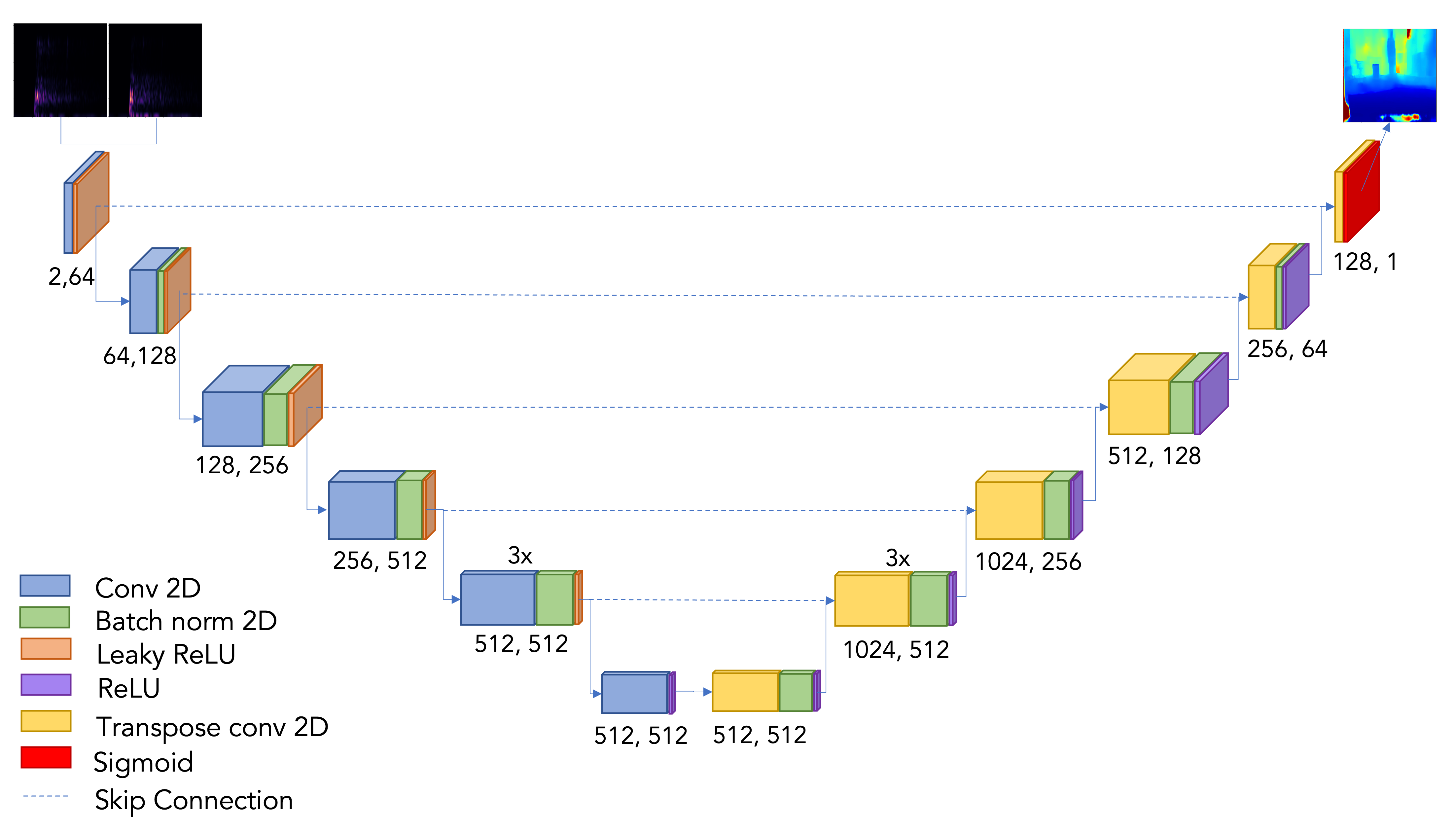}
    \caption{U-Net Architecture. Input spectrograms are processed to depth maps. Skip-connections used show better performance even though the spatial correspondence between input and output features is not clear.}
    \label{fig:UnetArchi}
    \vspace{-0.6cm}
\end{figure}

\textbf{U-Net Baseline.}
The extend to which audio can complement visual sensor data is not clear. Related work showed repeatedly the quality of the visual signal strongly dominates task performance. The compared work \autocite{parida2021beyond} investigates material-based attention to let the network choose the influence of each modality more explicitly. We investigate the claim that audio can help in bad visual conditions with a baseline, based on the audio signal alone. This serves to investigate the contribution of the audio signal in isolation with clear quality attribution. However, in all practical applications, the visual signal should be used as well.

We train a U-Net from audio only, similar to related work \cite{christensen2020batvision, gao2020visualechoes} and isolated on BV2 and BV1. We achieve solid results, correctly predicting free space, obstacles and the general room layout (see \Cref{fig:UnetResults}). Without further tweaks such as GCC-Phat features or using a GAN \autocite{christensen2020batvision,christensen2020gccbatvision}, the performance shows the exploitable quality of the data itself.

\textbf{Implementation Details.}
The only input are spectrograms, generated from waveforms with 512 frequency bins (nfft), 64 window and 16 hop length, resized to 256x256.

For BV2, best performance is obtained with depth clipped to 30m and audio cut accordingly. This can be explained as audio energy decays with distance so far traveling echoes are hard to distinguish from noise. For BV1, depth is clipped to 12m and the audio accordingly. Ground truth depth is always normalized using the chosen max depth value.

We use an 8 block U-Net with skip connection. Each encoder block is composed of 2D convolutions, batch normalization and leaky ReLUs. Each decoder block is composed of 2D transposed convolution, batch normalization and ReLUs (see \Cref{fig:UnetArchi}). We found skip-connections improve performance though contrary to a segmentation task their contribution is not yet clearly understood.
We train with 256 batch size, 0.002 learning rate for BV2 and 0.001 learning rate for BV1 with the AdamW \autocite{AdamWRef} optimizer and L1 loss.


\textbf{Results on Depth Prediction.}
With this simple baseline we retrieve the general geometry of the space (see \Cref{tab:bitd_table} and \Cref{fig:UnetResults}). 
On BV2, complex obstacles such as chairs are visible. When the visual sensors fails (e.g on glass), audio gives correct information about the depth.
The network generalizes well between different acoustic environment. It reproduces corridors robustly built with various material (e.g carpet floor and glass wall, stone and tiled floor, see \Cref{fig:UnetResults}). Outdoors the performance is diminished with the network underestimating depth systematically. This shows some inevitable bias of the data having a majority of indoor data.
Trained on BV1 data, corridors and obstacles are well reconstructed even though finer structures are equally lost.


\section{Conclusion}

Recent success in using simulated audio-visual data for scene understanding shows the potential of the audio modality. Depth prediction from sound alone or in addition to vision is possible, allowing to perceive the environment like a bat. The BatVision dataset will support the research community with large-scale real audio-visual data to improve task performance and uncover novel uses. 
We present an audio-only depth-prediction baseline as starting point and obtain good results when training a state-of-the-art approach on our data. Future datasets could include ultrasound \textit{chirps} to enable inaudible human-robot collaboration. 

\section{Acknowledgment}
We thank all collaborators: Daniel Lin for data collection, Jesper Haahr Christensen for insights into sonar and collaboration on the original idea, Karl Zipser for conception of the robot for BV1 and David Mazouz and Jacky Lech for advice and help building the robot for BV2. We acknowledge the support of the French Agence Nationale de la Recherche (ANR), under grant ANR-22-CE94-0003 and the US National Science Foundation (NSF), under grant 2215542.
\FloatBarrier

\printbibliography

\end{document}